\documentclass[10pt,a4paper,twoside,american]{article}
\usepackage{geometry}
\usepackage[toc,page,titletoc]{appendix}
\usepackage{authblk}
\usepackage{calc}
\usepackage{graphicx}
\usepackage[table]{xcolor}
\definecolor{shadecolor}{gray}{0.9}
\usepackage{natbib} 
\usepackage[font=small,labelfont=bf]{caption}
\usepackage{amsmath}
\usepackage{amssymb}
\usepackage{mathtools}
\usepackage{bm}
\usepackage[font=small,labelfont=bf]{caption}
\usepackage{subcaption}
\usepackage{algorithmic}
\usepackage{lineno}
\usepackage{multirow}
\usepackage{tabularx}
\usepackage[american]{babel}
\usepackage{amsfonts}

\usepackage[
breaklinks=true,
colorlinks=true,
linkcolor=gray,citecolor=gray,urlcolor=gray,filecolor=gray,
pdffitwindow,
bookmarks=true,
bookmarksopen=true,
bookmarksnumbered=true,
pdftitle={QS4D: Quantization-aware training for efficient hardware deployment of structured state-space sequential models},
pdfauthor={XXX},
pdfsubject={},
pdfkeywords={}
]
{hyperref}

\usepackage{cleveref}
\crefname{supp}{supplement}{Supplements}
\crefname{app}{appendix}{Appendices}
\crefformat{equation}{(#2#1#3)}
\crefformat{figure}{Figure\,#2#1#3}
\crefformat{section}{Sec.\,#2#1#3}
\crefformat{appendix}{App.\,#2#1#3}
\crefformat{table}{Table\,#2#1#3}

\usepackage{changes} 
\definecolor{YB}{RGB}{0,150,255}
\definecolor{TT}{RGB}{0,200,0}
\definecolor{DW}{RGB}{200,0,200}
\definecolor{SIS}{RGB}{100,0,100}
\definechangesauthor[color=YB]{YB}
\definechangesauthor[color=TT]{TT}
\definechangesauthor[color=DW]{DW}
\definechangesauthor[color=SIS]{SIS}

\usepackage{todonotes}

\title{QS4D: Quantization-aware training for efficient hardware deployment of structured state-space sequential models}

\def\shorttitle{QS4D}

\author[1]{Sebastian Siegel}
\author[1]{Ming-Jay Yang}
\author[2]{Younes Bouhadjar}
\author[2, 3]{Maxime Fabre}
\author[2, 4]{Emre Neftci}
\author[1, 4]{John Paul Strachan}

\affil[1]{\footnotesize%
  Peter Gr\"unberg Institute (PGI-14), Forschungszentrum J\"ulich GmbH and JARA, J\"ulich, Germany}
\affil[2]{\footnotesize%
  Peter Gr\"unberg Institute (PGI-15), Forschungszentrum J\"ulich GmbH and JARA, J\"ulich, Germany}
  \affil[3]{University of Groningen, Groningen, The Netherlands}
\affil[4]{Faculty of Electrical Engineering, RWTH Aachen University, Aachen, Germany}

\date{\footnotesize\today}

\usepackage{fancyhdr}
\title{QS4D: Quantization-aware training for efficient hardware deployment of structured state-space sequential models}

\begin{document}

\maketitle

\begin{abstract}

 Structured State Space models (SSM) have recently emerged as a new class of deep learning models, particularly well-suited for processing long sequences. Their constant memory footprint, in contrast to the linearly scaling memory demands of Transformers, makes them attractive candidates for deployment on resource-constrained edge-computing devices. While recent works have explored the effect of quantization-aware training (QAT) on SSMs, they typically do not address its implications for specialized edge hardware, for example, analog in-memory computing (AIMC) chips. 
 In this work, we demonstrate that QAT can significantly reduce the complexity of SSMs by up to two orders of magnitude across various performance metrics. We analyze the relation between model size and numerical precision, and show that QAT enhances robustness to analog noise and enables structural pruning. Finally, we integrate these techniques to deploy SSMs on a memristive analog in-memory computing substrate and highlight the resulting benefits in terms of computational efficiency.

\end{abstract}

\section{Introduction}
\begin{figure}[h]
    \centering
    \includegraphics{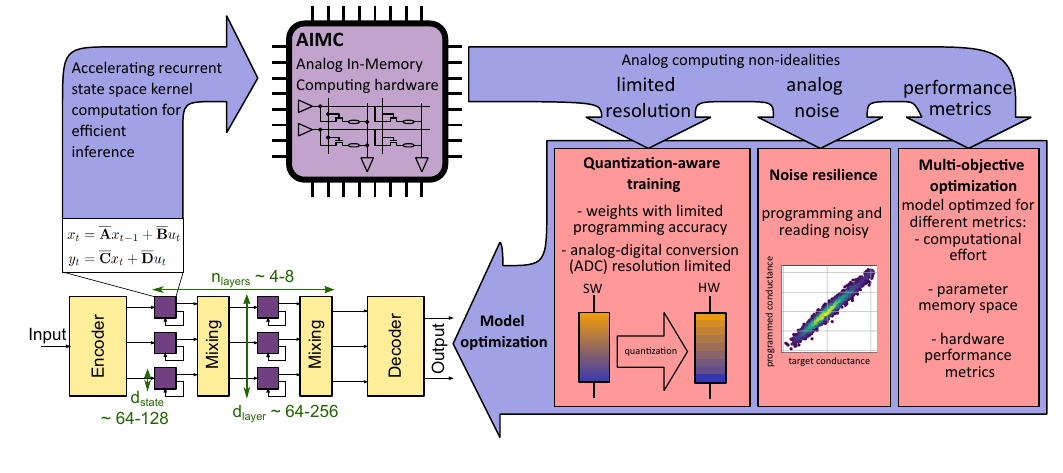}
    \caption{\textbf{Hardware-software co-design for efficient deployment of state-space models} Deployment of the recurrent state update and kernel function of the SSM on analog in-memory computing (AIMC) hardware offers acceleration potential in terms of energy and latency. Therefore, the non-idealities of AIMC, like limited resolution and transient noise, must be incorporated in the model.}
    \label{fig:top}
\end{figure}

Structured state space models (SSMs) have recently emerged as memory-efficient and effectively trainable neural networks for processing sequential data with long contexts \citep{gu2021efficiently, smith2022simplified, voelker2019legendre, gu2023mamba}. These models feature an expanded linear recurrent state space, decomposed into parallel kernels, which are projected into a lower-dimensional space before a nonlinearity is applied \citep{gu2020hippo}.
The linear nature of the recurrence enables fast training via convolutional operations on GPUs. The mathematically derived initialization schemes further facilitate the learning of long sequences. A layer is formed by multiple such kernels in parallel (see Figure \ref{fig:top}), hence the term ``structured''.
While sharing some conceptual similarities with Recurrent Neural Networks (RNNs), SSMs have been shown to outperform both traditional RNNs and even recent Transformer-based models on benchmarks involving long-context sequence modeling \citep{gu2022s4d, dao2024transformers}.
This makes SSMs a promising foundation for future Large Language Models (LLMs). Their recurrent state update mechanism allows them to encode historical information using a fixed amount of memory, in contrast to Transformers \citep{vaswani2017attention}, which require explicit storage of past tokens and incur a potentially unbounded memory cost as context length grows. Beyond LLMs, efficient sequence modeling is critical in many real-world applications, such as processing physiological signals in biomedical devices \citep{ballinger2018deepheart, ay2019automated, esteva2019guide}, or enabling long-term planning and perception in autonomous vehicles \citep{pandharipande2023sensing}. 
These applications often impose strong constraints on the power budget or allowed latency of the underlying hardware, which demands thorough hardware-software co-design of the algorithm and the underlying hardware. This can, on the one hand, guide hardware development dedicated to SSMs and for specific workloads, and on the other hand, is a tool to enable the execution of computationally demanding models on given resource-constrained hardware.
\par
A widely used strategy for deploying large machine learning models on edge hardware is quantization, which reduces memory footprint and computational load by lowering the precision of weights and activation signals. Prior work has shown that quantization-aware training (QTA) can effectively reduce the bit precision of key components in classical transformers, in some cases down to ternary values \citep{ma2024era}.
Recurrent neural networks, such as SSMs, however, are known for their high sensitivity to quantization \citep{ott2016recurrent}. Despite this challenge, quantization techniques have been investigated for SSMs, including S5 \citep{abreu2024q}, Mamba \citep{chiang2024quamba, tang2024bi}, and the related LMU \citep{blouw2020hardware}. However, except for the LMU work, these efforts primarily target implementations on general-purpose GPUs rather than specialized hardware.
More recently, a number of studies have demonstrated aggressive quantization techniques for the S4D model, highlighting the importance of this class of SSM, especially for edge-hardware deployment. For instance, the work by \cite{meyer2024diagonal} demonstrates the recurrent-mode deployment of an S4D model onto the Loihi neuromorphic processor \citep{davies2018loihi} and uses fine-tuning after post-training quantization. \cite{zhao2025quantizing} investigate quantization-aware fine tuning of small S4D models and find the recurrent kernel and the state to be the most sensitive parameters of the S4D model. Their analysis of the memory footprint already hints at the benefits of quantization on memory-constrained edge systems. \cite{abreu2024q} reach a similar conclusion for the memory footprint of the closely related S5 model.
However, key aspects of SSM deployment on edge-computing hardware remain underexplored, in particular the reduction in required hardware performance like compute performance and memory, the system's resilience to analog noise, and the potential for hardware-software co-design like size optimization and pruning.
\par  
In this study, we showcase the benefits of aggressive quantization-aware-training (QAT) across several metrics critical to edge-hardware devices, including memory usage and computational resources. As part of this analysis, we introduce a dedicated complexity metric tailored to analog in-memory computing systems, where computational resources are often dominated by the peripheral mixed-signal circuitry, particularly in architectures like memristive crossbar arrays \citep{xia2019memristive, cai2019mac}.
We further demonstrate that model size and precision can be effectively traded off, and that, specifically in analog in-memory computing systems, QAT enhances noise resilience.
Motivated by these findings, we showcase the deployment of SSM kernels on memristive crossbar arrays.
Memristive crossbar arrays are well-suited for this purpose, as they enable efficient vector matrix multiplication through analog computation. These arrays are matrix structures composed of resistive switching devices, an emerging type of non-volatile memory \citep{waser2007nanoionics}. Matrix elements are stored as conductances, inputs are applied as voltage vectors, and results are read out as currents. Leveraging Ohm’s and Kirchhoff’s laws, these arrays can compute vector-matrix multiplications in a single time step—or even continuously in time. However, due to their analog nature, they are inherently prone to noise, as illustrated in Figure~\ref{fig:top}.
While we investigated the impact and mitigation of static write noise in a previous study \cite{siegel2025imssa}, we here focus on transient read noise.
There are various memristive technologies under investigation.
In our case, we are using tantalum-oxide-based resistive switching devices, co-integrated with conventional $\text{180nm}$ CMOS circuits.
We thereby showcase the complete hardware-software co-design process of SSMs from QAT to deployment on an analog in-memory computing substrate.

\section{Results}
\subsection{Implementing quantization for S4D models}
\label{sec:qat}
S4D models are highly heterogeneous, combining recurrent kernels, nonlinear activations, and linear feed-forward layers. In this study, we focus on the S4D model with a diagonalized $\mathbf{A}$ matrix, which has become the standard approach, with respect to the original S4 model featuring a fully populated transition matrix per kernel \citep{gu2021efficiently}. A study on the quantization of the original S4 model can be found in the Supplementary Material.\par
To investigate the quantization opportunities of the S4D model, all parameters of the model's core, i.e., the diagonal $A$ matrix and $C$ vectors, the vector of trainable time steps $\Delta$ and the linear encoder and decoder layers are quantized by projecting them into an equally spaced ladder of integer values representing the maximum range of the respective parameter, described in detail in chapter ~\ref{chap:quant_methods}. Furthermore, quantization of the activations $y$, which are passed between layers, is also investigated as this communication requires a significant memory bandwidth that can be optimized.\par
When using the convolutional approach, which allows for fast training on GPU, the state $x$ is never explicitly computed. We thus proceed with an indirect quantization of the state. We use the convolutional kernel that is materialized for training, and the input to simulate a quantization of the state. Due to the convolution operation, the maximum resolution of the state is the sum of the precision of the kernel and the input. Therefore, if the state is quantized to $k$ levels, the quantization of the input and the convolutional kernel are set to $k/2$. However, it should be noted that this does not yield the exact same numerical results as the recurrent computation. \cite{zhao2025quantizing} study the direct quantization in the recurrent mode.
%
\subsection{Quantization-aware training of SSMs allows aggressive quantization}
\begin{figure}[h]
    \centering
    \includegraphics{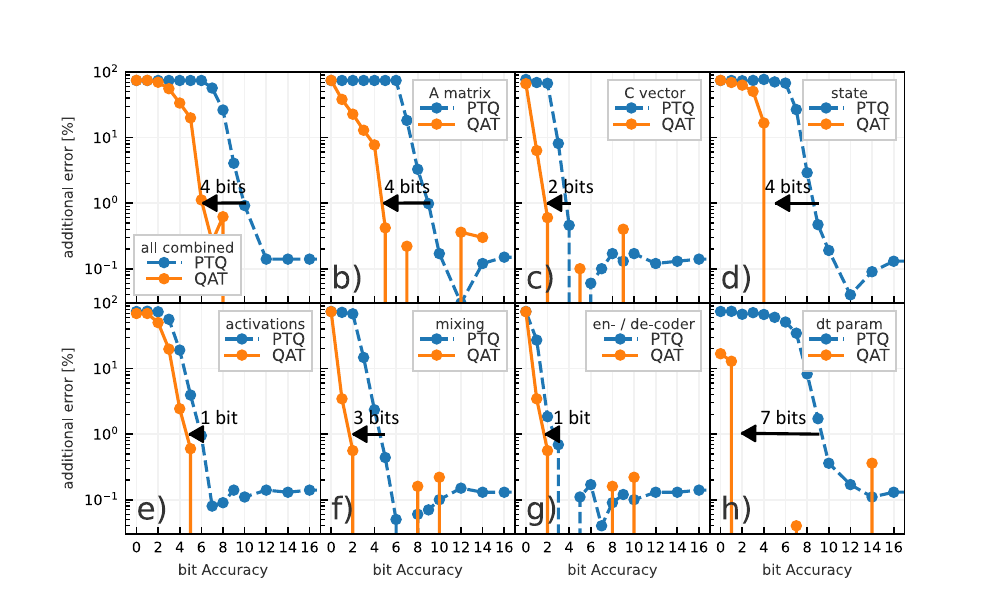}
    \caption{\textbf{Quantization-aware training of sequential CIFAR10 (grayscale).} a) For homogeneous quantization, QAT allows a reduction to 6 bits for all model parameters before the error increases above $1\%$ over the base line. For PTQ, this already happens at 10 bits accuracy. b) - h) For the quantization of only individual parameters, a varying benefit of QAT over PTQ can be observed. The gain in quantization is especially high for the recurrent kernel parameters (b), d), and h)).}
    \label{fig:S4D_cifar10}
\end{figure}
The plots in Figure \ref{fig:S4D_cifar10} show the impact of quantization of all key parameters of the model for the sequential gray-scale CIFAR10 benchmark, evaluating the model's capacity to remember long sequences. We display the additional prediction error over a fully floating-point trained baseline when applying a certain quantization to the parameter or a combination thereof. Starting from the right side with 16 bit parameter accuracy, a more aggressive quantization is applied going the the left side of the plot. The results for post-training quantization (PTQ) are given in blue (dashed line) while quantization-aware training (QAT) is shown in orange (solid line). \par
When quantizing all parameters homogeneously (Figure \ref{fig:S4D_cifar10} a)) it can be seen that PTQ leads to an increased error starting from 12 bit accuracy to lower resolutions. At 10 bit, the additional error reaches 1 \%, which is used as a level for comparison here. With QAT, in contrast, a homogeneous quantization of 6 bit can be reached before crossing this threshold. \par
In fact, we observe this benefit of QAT over PTQ for all parameters. It is especially striking for parameters partaking in the recurrent state update like the transition matrix $\mathbf{A}$ (Figure \ref{fig:S4D_cifar10} b)), the state (Figure \ref{fig:S4D_cifar10} d)), and the time-step $dt$ (Figure \ref{fig:S4D_cifar10} h)).

\begin{table}[h]
\centering
\begin{tabular}{c||c||c||c|c|c|c|c|c|c|c|c}

     & baseline & & all & A & B & C &state & activations & mixing & en-/decoder & $\Delta \text{t}$ \\
     \hline
     \hline
     & & PTQ & 10 & 9 & NA & 4 & 9 & 6 & 5 & 3 & 10 \\
    sCIFAR10 & 86.5\% & QAT & 6 & 5 & NA & 2 & 5 & 5 & 3 & 2 & 2\\
     & & gain & 4 & 4 & NA & 2 & 4 & 1 & 2 & 1 & 8 \\
     \hline
     & & PTQ & 16 & 13 & 2 & 2 & 7 & 5 & 2 & 2 & 16\\
    Pathfinder & 97.6\% & QAT & 5 & 3 & 2 & 2 & 5 & 4 & 2 & 2 & 1\\
     & & gain & 11 & 10 & 0 & 0 & 2 & 1 & 0 & 0 & 15 \\
     \hline
     & & PTQ & 12 & 6 & NA & 6 & 12 & 8 & 5 & 3 & 7 \\
    HD & 100.0\% & QAT & 5 & 2 & NA & 1 & 4 & 3 & 1 & 1 & 1 \\
     & & gain & 7 & 4 & NA & 5 & 8 & 5 & 4 & 2 & 6\\

\end{tabular}
     \caption{\textbf{Maximum quantization to stay below 1$\%$ of additional error with PTQ and QAT for the sequential CIFAR10, Pathfinder, and Heidelberg Digits (HD) tasks, in bit.} "all" denotes the values for homogeneous quantization of all parameters simultaneously. For the other columns only the indicated parameter is quantized.}
     \label{tab:quantization}
\end{table}
\subsubsection*{Pathfinder dataset}
The impact of quantization on the model performance depends on the task. For comparison, we investigate the Pathfinder task, another benchmark task from the Long Range Arena suite \cite{tay2020long} like sCIFAR, which also requires to memorize a long sequence. Table \ref{tab:quantization} lists the maximum quantizations achievable to stay below $1\%$ of additional error for both PTQ and QAT. On the Pathfinder dataset, QAT allows for a 5 bit homogeneous quantization of the whole model while PTQ fails below 16 bit. This is a gain of 11 bit in parameter size. The largest gains for the individual parameters can be found for the A matrix and the $dt$ parameter. \par
Also for this challenging benchmark task, QAT leads to significant gains over PTQ, especially in the recurrence matrix A, which also has a large impact on the computational complexity.

\subsubsection*{Keyword spotting (Heidelberg digits (HD))}
A possible application for state space models in edge computing systems is the classification of temporal data like audio streams, for example, in a keyword spotting application. This task is significantly different from the synthetic ones in the Long Range Arena suite that focus mostly on the system's capability for memorization. We thus investigate the quantization behavior for the classification of the Heidelberg Raw Audio Digits \cite{cramer2020heidelberg}. The dataset consists of recordings of English and German spoken digits from zero to nine of different speakers. In this work, we limit the dataset to the English subset.\par
Like for the benchmark tasks, for the HD dataset, a large gain of 7 bits can be achieved with QAT over PTQ. However, when discriminating by parameter, the major gains can be found for the recurrence matrix A, the state, and the activations. For CIFAR10 and Pathfinder, only minor gains can be reached for the latter parameter. 

\subsection{Efficiency gains through aggressive quantization}
\begin{figure}[h]
	\centering
	\includegraphics{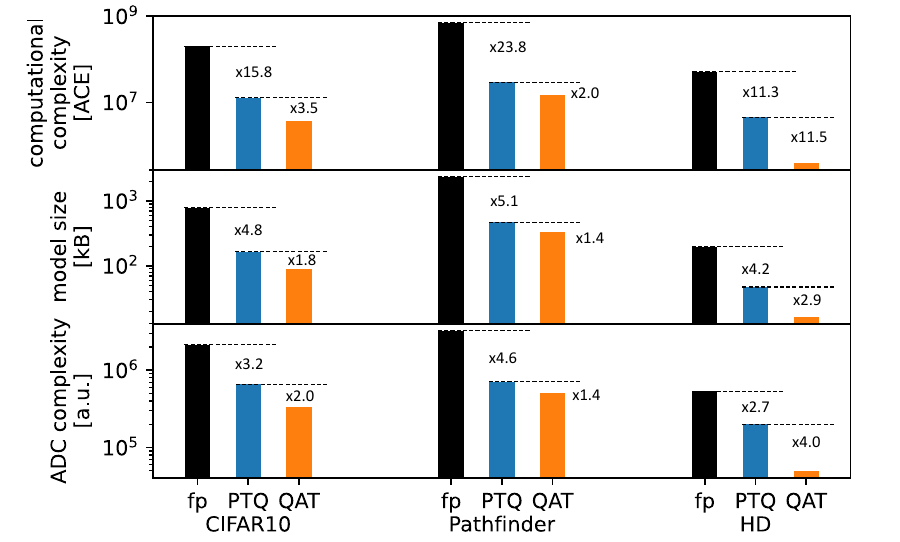}
	\caption{\textbf{Quantization benefits in different hardware metrics} The graphs show the respective metrics for models with maximum quantization. a) The Accumulated Computational Complexity (ACE) for all investigated tasks decreases by at least a factor of 11 by quantizing the model after training. QAT allows for an additional decrease by a factor of 2 to 11.5. b) The memory required to store all models parameters decreases with PTQ by at least a factor of 4.2 and an additional 1.4 to 2.9 with QAT. c) the Analog-Digital Conversion metric estimates the peripheral complexity in an AIMC hardware implementation. Also here benefits due to PTQ are surpassed by QAT.}
	\label{fig:ACE}
\end{figure}
The effect of quantization on the efficiency of hardware implementations can be quantified by different metrics, depending on the type of hardware substrate. For conventional digital processing units, such as GPUs, the number of Multiply \& Accumulate (MAC) operations is important. For memory or memory bandwidth-constrained systems like microcontrollers, the model size in terms of storage for the model parameters can become an important factor. To evaluate the computational complexity, we employ the ACE metric \citep{zhang2022pokebnn} as outlined in Section \ref{sec:ACE}. For the memory footprint, we accumulate the bit values for all parameters (see Section \ref{sec:modelsize}). Finally, for analog in-memory computing (AIMC) solutions, the peripheral circuitry for the analog-to-digital conversion (ADC) after the analog vector-matrix-multiplication is often crucial in terms of chip area, latency, and energy consumption. We propose to estimate this complexity by accumulating the precision bits of state and activation read-outs. For details, we direct the reader to Section \ref{sec:adc}).\par
Using the quantization levels obtained above, significant reductions in all of these metrics can be obtained. Figure \ref{fig:ACE} shows these results for reference floating-point (fp) models, and models with maximum quantization below $1\%$ additional error for the PTQ and QAT methods. We cover sequential CIFAR10, Pathfinder, and the HD audio dataset. By PTQ, the computational effort decreases already by a factor of 11.3 to 23.8 (Figure \ref{fig:ACE} a)) which is further surpassed by the QAT models by additional factors between 2 and 11.5. Similarly, the model's size in Figure \ref{fig:ACE} b) benefits by at least a factor of 4 from PTQ and reduces further down by a factor up to 2.9 with QAT. In the same manner, the peripheral ADC complexity is decreased by a factor of 2.7 to 4.6 by PTQ alone and by another factor of 1.4 to 4. For all metrics, the largest benefit of QAT over PTQ can be found for the HD audio classification dataset, which is one of the most likely application areas for an SSM model deployed on edge hardware.

\subsection{Trading model size for quantization}
Model dimensions, especially its width, are another central element on top of quantization to play with a model's efficiency-accuracy trade-off for edge implementations.  In the following, we investigate the impact of the model size in relation with quantization, also trying to establish a sweet spot for accuracy and efficiency between the two. An educated choice of model size and resolution can significantly impact performance for the deployment of the model to a given hardware substrate with memory or computational constraints.\par
\begin{figure}[h]
    \centering
    \includegraphics{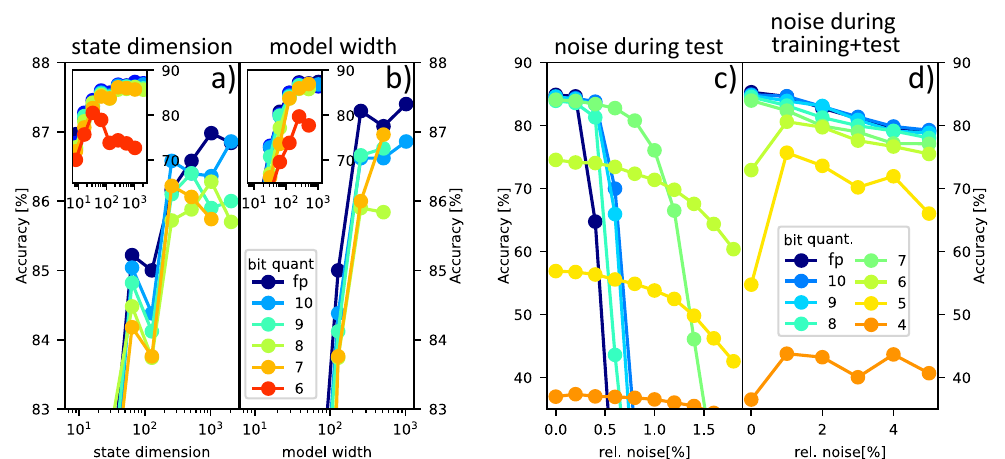}
    \caption{\textbf{Relation of model size and quantization} a) Below a state size of $\mathbf{N} = 256$, an increase of the state dimension increases the model accuracy and can compensate the loss due to quantization. For higher state dimension the accuracy plateaus. b) A similar behavior can be observed for the model width. \textbf{Noise resilience} c) Increasing transient noise in the A matrix causes a degradation of the performance. Aggressive quantization can slow down this degradation, but can also reduce the performance in general. d) Introducing transient noise during training makes all models more resilient against noise and can even increase the accuracy over training without noise. All trainings are performed on the sequential gray-scale CIFAR10 benchmark task.}
    \label{fig:noisesize}
\end{figure}

There are three parameters determining the size of the S4D model: the state dimension $\mathbf{N}$, the model width $\mathbf{H}$, and the number of layers. The latter is very specific for each task and is therefore fixed in this study. Figure \ref{fig:noisesize} a) and b) depict the task performance for the sequential CIFAR10 task for different model sizes and full-model quantizations. Missing data points are due to the training failing with the given parameters. \par
In Figure \ref{fig:noisesize} a), we vary the state dimension on the x-axis. It can be seen that for most investigated homogeneous quantization levels, the model performance increases with increasing state dimension up to a maximum, which is dependent on the quantization. For low resolution, however, the performance decreases for larger state dimensions, as can be seen for 6 bits in the inset. \par
For the variation of the model width in Figure \ref{fig:noisesize} b), we observe similar behavior. The performance increases with the width until a saturation is reached at $\mathbf{H} = 128$, at an accuracy value highly correlated with the quantization level.\par
Another way to reduce the model size is by pruning. Due to their highly structured shape, S4D models can significantly benefit from structural pruning if complete kernels can be eliminated. This applies especially for an in-memory hardware implementation, because kernels and states need to be materialized in hardware. When investigating the prunability of a model with homogeneous quantizations on the example of the CIFAR10 benchmark task, for each layer, a different number of complete kernels can be pruned (see Table \ref{tab:prune}). For the model with 7 bit homogeneous quantization, in the first layer, more than half of the kernels can be eliminated without losing more than 1\% task performance, while in the third layer, only about 5.49\% kernels can be pruned. When comparing models with different homogeneous quantization (here 7 bit, 10 bit, and floating-point), a more aggressive quantization allows for more aggressive structural pruning across all layers. \par
The differences in pruning potential arise especially for the structural pruning of complete kernels. While there are also small benefits of highly quantized models for unstructured kernel pruning, for the unstructured pruning of the linear layers, no significant difference can be observed.

\begin{table}[h]
    \centering
    \begin{tabular}{c|c|c||c||c}
        Layer & & 7 bit & 10 bit & fp \\ \hline
        
         & kernel total & 62.89\% & 58.79\% & 60.35\% \\ \hline
        1 & kernel structural & 53.125\% & 42.188\% & 39.063\% \\ \hline
         & linear layer  & 61.23\% & 65.24\% & 78.63\% \\ \hline \hline

         & kernel total & 22.46\% & 6.64\% & 6.05\% \\ \hline
        2 & kernel structural & 14.848\% & 3.125\% & 0.0\% \\ \hline
         & linear layer  & 35.57\% & 35.06\% & 47.95\% \\ \hline \hline

         & kernel total & 9.77\% & 7.23\% & 6.64\% \\ \hline
        3 & kernel structural & 5.469\% & 0.0\% & 0.0\% \\ \hline
         & linear layer  & 39.02\% & 37.86\% & 43.23\% \\ \hline \hline

         & kernel total & 16.21\% & 10.94\% & 4.88\% \\ \hline
        4 & kernel structural & 7.031\% & 0.781\% & 0.0\% \\ \hline
         & linear layer  & 50.66\% & 48.82\% & 49.57\% \\
    \end{tabular}
    \caption{\textbf{Post training pruning}}
    \label{tab:prune}
\end{table}

%

\subsection{Quantization for noise robustness}
A possible way to efficiently implement the recurrent kernels of S4 models is with crossbar arrays of non-volatile analog memory devices in an in-memory compute fashion. This significantly accelerates the recurrent vector-matrix multiplication, but it also introduces noise in the computation. One crucial noise observed in analog storage devices is electronic read noise on the conductance which would reflect as transient noise on the kernel parameters. \par
A deployment on analog in-memory compute substrates thus necessitates a study of such noise on the task performance. Therefore, we subject the kernel parameters of models trained with QAT for different quantizations with Gaussian transient noise during the inference, after training without. In Figure\ref{fig:noisesize} c) the model performance for an increasing relative parameter noise is depicted for different quantization levels for the CIFAR10 task. It can be seen that the model with full precision (fp) and models trained with high accuracy parameters (10 bit to 8 bit) lose performance quickly while increasing noise. For models trained with more aggressive quantization, the performance drops slower with increasing noise. The highest quantized models however exhibit a lower performance from the start but can surpass less quantized models for high noise levels. \par
We suggest that this acquired noise robustness stems from the fact that quantization itself is a form of noise. QAT thus constitutes a form of training with noise. To prove this, we explicitly introduce Gaussian noise to the recurrent kernel parameters during training. The results can be seen in Figure \ref{fig:noisesize} d) (the large range of the x-axis should be noted when comparing to c)). All models, especially those with high precision parameters, show a slower decrease in task performance with increasing noise levels, when they are trained with the respective noise. For highly quantized models (6 bit, 5 bit, and 4 bit) adding noise during training even increases their performance and restores some of the losses due to the quantization. \par
In conclusion, aggressive quantization through QAT can make a model more resilient to parameter noise. For a known level of noise however, training with this noise still yields superior results.

\section{Implementation of S4D kernel on memristive crossbar arrays}
To enable efficient execution on GPUs, the S4D model’s kernel is unrolled into a convolutional kernel, and the input is processed in batches. This allows the GPU to leverage highly parallel matrix-matrix multiplication. However, implementing this approach on dedicated hardware requires storing the unrolled convolutional matrix and input data across multiple time steps, which is impractical for resource-constrained edge systems due to memory limitations. A more viable alternative for such systems is a streaming implementation, where the input signal is processed sequentially, one time step at a time. This approach necessitates implementing the original state-space formulation, as given in Equation (\ref{eq:kernel_discrete}), directly in hardware. It relies on fast and efficient vector-matrix multiplication, while minimizing memory usage to the storage of latent states only.\par
Memristive crossbar arrays (mCBAs) have been proposed in the literature as a means to enable time- and energy-efficient vector-matrix multiplication within a single time step. In this work, due to system constraints, we implement the time-discrete version of the S4D model. However, in principle, mCBAs can also support time-continuous operation and thus could be leveraged to implement the continuous-time formulation of the model.\par
\begin{figure}[h]
	\centering
	\includegraphics{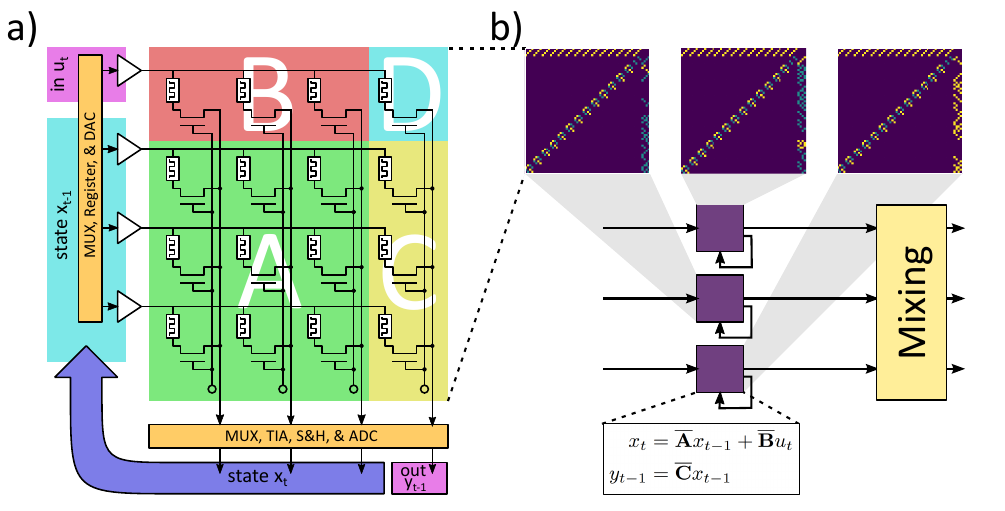}
	\caption{\textbf{In-Memory State-Space Model Accelerator (IMSSA) kernel implementation on an mCBA.}}
	\label{fig:MESSMA_kernel}
\end{figure}

To minimize data movement between compute blocks, as much computation as possible should be performed within the same crossbar array. To this end, we proposed the In-Memory State Space Accelerator (IMSSA) architecture \cite{siegel2025imssa}, which maps the S4D kernel computation such that it performs not only the $\mathbf{\overline{A}}x$ operation, but also the $\mathbf{B}u$ and $\mathbf{C}x$. The latter two are vector-scalar operations and thus require only a single row or column within the crossbar array, as illustrated in Figure \ref{fig:MESSMA_kernel}. While this reduces the space available for the $\mathbf{A}$ matrix, it enables the entire S4D kernel to be computed within a single mCBA. The state is updated by accumulating the resulting current vector at each time step. As a consequence, the $\mathbf{C}$ vector is not multiplied directly with the current state $x_t$, but rather with a one-time-step delayed version. The IMSSA kernel computation is therefore given by:
\begin{equation}
    \begin{aligned}
        x_{t} &= \overline{\mathbf{A}}x_{t-1} + \overline{\mathbf{B}}u_t\\
        y_{t-1} &= \overline{\mathbf{C}}x_{t-1} + \overline{\mathbf{D}}u_{t}
    \end{aligned}
    \label{eq:kernel_discrete_mem}
\end{equation}
which is a small deviation from the original kernel in Equation (\ref{eq:kernel_discrete}).
\begin{table}[h]
\centering
\begin{tabular}{c||c|c|c|c|c|c|c}
    parameter & $n_{layers}$ & $\mathbf{N}$ & $\mathbf{H}$ & $r_{state}$ / $r_{act}$ & $r_{kernel weights}$ & SW accuracy & mcba accuracy\\
    \hline 
    value & 1 & 14 & 3 & 8bit & 4bit & 95.8\% & 95.3\%\\

\end{tabular}
\caption{Parameters used for small (sub-dataset) of Keyword spotting (HD)}
\label{tab:hdsmall_params}
\end{table}
The proposed mCBA system comprises three memristive crossbar arrays, each configured as a 64×64 matrix of memristive devices. Further details on the system architecture are provided in the Supplementary Information. A performance comparison with a commercial edge GPU (NVIDIA Jetson Nano) is presented in Table \ref{tab:draglift1}. Representing a single complex matrix element requires a 4×4 submatrix of memristive devices, as detailed in the Methods section \ref{sec:mcba}. Consequently, each mCBA can represent a 16×16 matrix of complex values. One row and one column are reserved for the $\mathbf{B}$ and $\mathbf{C}$ vectors, respectively. We allocate a 14×14 matrix for $\mathbf{A}$ and assume a device programming precision of 3 bits per cell.
Using these constraints, we train an S4D model—configured according to the parameters listed in Table \ref{tab:hdsmall_params}—on a two-class subset of the HD audio recognition dataset, achieving a test accuracy of 95.8\%. The resulting kernels are programmed on the mCBA system and read out to achieve realistic write and read noise distributions. With these programmed kernels, the system achieves classification accuracies ranging from 83.69\% to 86.41\%. The observed drop in accuracy is attributed to programming multiple parameters with differing dynamic ranges onto the same memristive substrate, which reduces their effective quantization resolution. With the method of a common constant maximum value for all kernel parameters outlined in \citep{siegel2025imssa}, we achieve an accuracy of 95.3\%, almost achieving software task performance.\par

%
\renewcommand{\arraystretch}{1.3}
\captionsetup[table]{position=bottom}
\begin{table}[ht]
\centering
\begin{tabular}{  c || c | c }
  \textbf{Benchmark metric} & \textbf{This work} & \textbf{Jetson Nano} \\ \hline \hline
  Technology & 180nm & 20nm  \\ \hline
  Area (mm2) & 136.85 & 118.0\\ \hline
  Bit Precision & up to int 4 & float 16 \\ \hline
  Throughput (GOPS) & 122.88 & 472 \\ \hline
  Power (W) & 16.47e-3 & 10 \\ \hline
  Energy efficiency (TOPS/W) & 7.46 & 47.2e-3 \\ \hline
  Area efficiency (TOPS/mm2) & 0.9e-3 & 4 \\
\end{tabular}
\caption{Comparison with digital CMOS AI inference accelerators per inference}
\label{tab:draglift1}
\end{table}
%
%
\section{Discussion}
In this work, we discuss the process of deploying a structured state-space sequential model (specifically, the S4D model) to edge-computing accelerators. The goal of this process is to accelerate the recurrent state update during inference, which is inefficient on classical GPU compute substrates.
\par
In a first step, we demonstrate how quantization-aware training can successfully be applied to S4D models. For various classes of sequence learning tasks, QAT allows for a more aggressive quantization than PTQ, especially for the recurrent kernel parameters. This is an important step, because edge-computing substrates often offer only compute elements with a limited precision. We find the largest benefits of QAT over PTQ for the recurrent update matrix $A$ and the state, which is confirmed by the parallel study of \cite{zhao2025quantizing} for small models and tasks. We here show the same trend for larger models and a wide range of applications, from perfect memory tasks like the Pathfinder challenge to audio classification of spoken digits.
\par
Depending on the task, aggressive quantization with QAT reduces the computational complexity by multiple orders of magnitude, the memory required for the model parameters by approximately a factor of 10, and also the complexity for analog-digital conversion in an AIMC substrate significantly.
Furthermore, QAT allows for aggressive structural pruning after training, further enhancing these gains. These findings are promising for the deployment of S4D models on edge-computing hardware, which is often constrained by computational power, memory, or peripheral complexity in the case of AIMC.
\par
Edge-computing substrates often come with restrictions on the size of matrix operations they can support efficiently. We demonstrate how the size of the recurrent kernels and the number of kernels per layer can be traded against the level of quantization that can be applied. A similar scaling law was observed for large language models (LLMs) \citep{kumar2024scaling}. This opens up the design space for hardware-software co-design for tailored accelerator systems and makes it possible to fit models to the specific properties of a given edge-computing system.
\par
Especially AIMC systems suffer from inherent noise when reading and writing parameters. After investigating the read noise in a previous study, we showcase here how QAT makes models more resilient against transient read noise. More aggressive quantization leads to more noise resilience but there is a trade of with the generally lower task performance at very high quantizations. For known noise levels, we demonstrate that training with noise can lead to even more resilient models and even increase the task performance of highly quantized models.
\par
Finally, we demonstrate the successful deployment of an S4D model for audio classification on a memristive AIMC substrate using the techniques showcased in this paper and the deployment methods described in \cite{siegel2025imssa}. For that, we use a mapping of the recurrent kernel to a memristive crossbar array that combines all kernel functions in one array and thus allows to perform state update and output to be computed in a single time step. \par
In this study, we showcase the benefits of QAT of SSMs for deployment on edge-computing substrates by reducing computational complexity and memory footprint. Specifically for AIMC we demonstrate how the two key factors peripheral circuit complexity and noise resilience benefit from QAT and conclude the study by showcasing the deployment of S4D kernels on memristive crossbar arrays. 
Thereby, we pave the way for efficient sequence processing with ultra-long contexts at the edge.

\section{Methods}
\subsection{Structured state space models}
As proposed in the work by \cite{gu2021efficiently}, each layer of the S4 state space model consists of a transition matrix $\mathbf{A} \in \mathbb{R}^{NxN}$ describing the evolution of the state $\text{x(t)} \in \mathbb{R}^N$ over time. The input $\text{u(t)} \in \mathbb{R}$ influence the state $\text{x(t)}$ via the matrix $\mathbf{B} \in \mathbb{R}^{Nx1}$ which additionally projects the input signal into the dimension of $\text{x(t)}$. The output $\text{y(t)}$ of a layer is a combination of a projection of the state $\text{x(t)}$ via the matrix $\mathbf{C} \in \mathbf{R}^N$ and the input signal $\text{u(t)}$ via the matrix $\mathbf{D}$. In this work, as is usually the case, $\mathbf{D}$ is set to zeros.

\begin{equation}
    \begin{aligned}
        \frac{d x(t)}{dt} &= \mathbf{A}x(t) + \mathbf{B}u(t) \\
        y(t) &= \mathbf{C}x(t) + \mathbf{D}u(t)
    \end{aligned}
    \label{eq:kernel_continuous}
\end{equation}

$\text{H}$ blocks of this structure are stacked to form an S4 layer, which is repeated $\textbf{n}_{layer}$ times to form the model. Initially, the input signal to the network is projected to the dimension $\textbf{H}$ of the model by a linear encoder. The output of the model is again projected via a linear transformation to the output dimension. In this work, classification tasks are investigated. Therefore, the output dimension is the number of classes. \par
For the execution on time-discrete compute systems like CPUs and GPUs, the time-continuous kernel of the S4 model needs to be discretized in time. In this work, a Zero-Order-Hold (ZOH) discretization is chosen as proposed in \cite{gu2021efficiently}. Given a time step $\Delta$, this yields the kernel formulation

\begin{equation}
    \begin{aligned}
        x_{t} &= \overline{\mathbf{A}}x_{t-1} + \overline{\mathbf{B}}u_t\\
        y_t &= \overline{\mathbf{C}}x_t + \overline{\mathbf{D}}u_t
    \end{aligned}
    \label{eq:kernel_discrete}
\end{equation}

The time step $\Delta$ is individual for each S4 block and becomes a trainable parameter. This yields an additional parameter vector $\Delta \in \mathbb{R}^N$ per layer. \par
To reduce the computational complexity of the state update, the $\mathbf{A}$ matrix is diagonalized. Since this is by default not possible, \cite{gu2022s4d} propose splitting $\mathbf{A}$ into a normal (and thus diagonalizable) part and a low-rank residual. Dropping this residual has only a minor impact on the task performance, but reduces the computational burden to a linearly scaling element-wise multiplication. These models are called S4D models and are used for the remainder of this analysis.\par
The model as given in \ref{eq:kernel_discrete} is a recurrent model. This would mean that the state $x$ is read and re-computed for each time step of a sequence and stored in memory after each step. This would lead to a high IO bandwidth and render the model inefficient on modern GPU systems. Therefore, \cite{gu2021efficiently} propose to take advantage of the linearity of \ref{eq:kernel_discrete} and unroll the recurrence into a convolutional kernel to execute multiple steps of \ref{eq:kernel_discrete} in a single matrix-matrix multiplication like 

\begin{equation}
	\begin{aligned}
		{K} &= (\overline{C}\overline{B}, \overline{C}\overline{A}\overline{B}, \overline{C}\overline{A}^k\overline{B}, ... , \overline{C}\overline{A}^L\overline{B})\\
		{y} &= {K} \star {u}
	\end{aligned}
	\label{eq:conv}
\end{equation}

for an $L$ time-step long sample. For practical implementation, this convolution is additionally transformed to the frequency space by fast-fourier-transformation and the convolution becomes a multiplication.\par
Gu et al. found that replacing $\mathbf{B}$ with an identity and not training it in many cases only leads to a small drop in performance \citep{gu2022s4d}. Since this would also increase the computational efficiency of the model, we apply this approach whenever possible. For this study, this means that the analysis on the CIFAR10 and the Heidelberg Digits data set employs a $\mathbf{B}=1$. For the training of the pathfinder data set, however, a trainable $\mathbf{B}$ matrix is necessary, as previous studies also showed.

\subsection{Quantization}
\subsubsection{Quantization methods}
\label{chap:quant_methods}
Quantization describes the resolution of weights and activations in the model, which often involves changing from floating-point precision to fixed-point precision, also called integer precision. However, the operations reported in this work are executed on GPUs, and for the sake of comparability, all are performed with 32-bit floating-point values for all parameters. Therefore, to emulate a lower resolution, parameters are subjected to artificial quantization by rounding each value within a weight layer or activation vector to the next value in an equally spaced grid, with the target quantization reflected in the number of grid cells. The range of the grid can either be determined by the maximum and minimum values or the average value in the respective layer. In this work, the grid spans around zero with a range of the maximum absolute value. Other works like \cite{ma2024era} also use the average of absolutes as a range. The formula for quantization can be expressed as
\begin{equation}
    \begin{aligned}
        f_{\text{scale}} &= \text{max}\left(abs\left(x\right)\right) \\
        x_{i, quant} &= round\left(x_i * \frac{n_{levels}}{f_{scale}}\right) * \frac{f_{scale}}{n_{levels}}
    \end{aligned}
    \label{eq:quantization}
\end{equation}

\subsubsection{Post-training quantization (PTQ)}
For the post-training quantization, the model is trained with all parameters and activations being 32 bit floating-point values. After the training, parameters are subjected to the quantization formula \ref{eq:quantization}. The same method can also be used to quantize the activations. In this specific case, this means that the signals propagated between layers ($y_t$ in \ref{eq:kernel_discrete}) are quantized. 

\subsubsection{Quantization-aware training (QAT)}
In contrast to PTQ, QAT includes quantization already in the training cycle. As described in \cite{ma2024era}, this is done in a way that the quantization method is applied during the forward pass, but not during the backward pass, also called Straight-through Estimator. This is achieved by detaching the quantization from the computational graph for the autograd function. 

\subsection{Hardware metrics}

\subsubsection{Computational complexity}
\label{sec:ACE}

To quantify the computational complexity, we employ the Arithmetic Computational Effort (ACE) metric \citep{zhang2022pokebnn}, which computes the computational effort in relation to single-bit MAC operations. For the kernel and linear layer operations of the S4D model, the ACE complexity can be computed as follows:

\begin{equation}
    \begin{aligned}
        ACE_{Ax} &= c * N * r_A * r_{act} * H * n_{layer}\\
        ACE_{Bu} &= c * N * r_B * r_{act} * H * n_{layer}\\
        ACE_{Cx} &= c * N * r_C * r_{act} * H * n_{layer}\\
        ACE_{linear} &= H * H * r_{act} * r_{linear} * n_{layer}\\
        ACE_{en-, decoder} &= H * (n_{in} + n_{out} * r_{act} * r_{coder}\\
    \end{aligned}
	\label{eq:ace}
\end{equation}

where $r_A$ is the bit resolution of the $A$ matrix,$r_B$ and $r_C$ is the bit resolution of the $B$ and $C$ vectors, $r_{act}$ is the bit resolution of the activation and the state, and $n_{layer}$ is the number of layers. $c$ is a constant factor that is $1$ for real values and $4$ for a complex values. $r_{linear}$ is the bit resolution of the linear mixing layers, while the resolution of the encoder and decoder layers is given by $r_{coder}$. $n_{in}$ and $n_{out}$ are the number of input and output channels of the model, respectively.
The sum of all contributions is the total ACE.\par

\subsubsection{Model size}
\label{sec:modelsize}
The model size is the sum of bits that its parameters occupy in memory. This comprises the kernel parameters, the linear mixing, encoder, and decoder layers. It can be calculated as

\begin{equation}
    \begin{aligned}
        Mem &= c * N * r_A * H * n_{layer}\\
        &+ c * N * r_B * H * n_{layer}\\
        &+ N * r_C * H * n_{layer}\\
        &+ H * H * r_{linear} * n_{layer}\\
        &+ H * (n_{in} + n_{out} * r_{coder}\\
    \end{aligned}
\end{equation}

where here $c$ is $1$ for real valued parameters and $2$ for complex.

\subsubsection{ADC complexity}
\label{sec:adc}
As a measure for the complexity of the analog-to-digital conversion circuitry necessary for an analog in-memory computing implementation, we propose the number of bits that need to be converted. This can be calculated as

\begin{equation}
    \begin{aligned}
        ADC_{kernel} &= (c * N * r_{state} + r_{act}) * H * n_{layers}\\
        ADC_{mixing} &= r_{act} * H * n_{layer}\\
        ADC_{coder} &= r_{act} * (H + n_{out})\\ 
    \end{aligned}
\end{equation}

\subsection{Memristive crossbar hardware implementation}

\subsubsection{In-memory S4D kernels}

Memristive devices are an emerging class of non-volatile memory technology. Their electrical conductance can be modulated by applying electrical stimuli, often in a gradual and controllable manner. Various physical mechanisms underlie this behavior. In this work, we employ transition metal oxide-based memristive devices \citep{SuperT} that utilize the Valence Change Mechanism (VCM) to alter their conductance. Specifically, the devices consist of a tantalum oxide (TaOx)/tantalum switching layer, sandwiched between two noble metal electrodes. Each memristive device is integrated in series with a conventional CMOS transistor, forming a 1-transistor-1-resistor (1T1R) configuration. These 1T1R cells are arranged in a matrix-like structure known as a memristive crossbar array (mCBA). Mathematical matrices can be encoded in the mCBA as conductance values $g$. When an input vector is applied as a voltage vector $v$, the resulting output current vector $i$ reflects the outcome of a vector-matrix multiplication. This operation is inherently parallel and can be executed in a single computational step, and even supports time-continuous processing under appropriate conditions.\par
The conductances of memristive devices are inherently positive and real-valued. To represent complex-valued matrices—such as the $\mathbf{A}$ matrix of an S4D model, which may also contain negative elements—each matrix element must be decomposed and encoded using multiple memristive conductances. A signed real value is typically implemented using two conductances, enabling differential encoding. Furthermore, due to technical constraints, memristive arrays often support only positive voltage inputs. As a result, the computation of a single matrix element—whether real or complex—must be expanded accordingly to account for these hardware limitations: 
\begin{equation}
    \begin{pmatrix}
        i^+\\
        i^-
    \end{pmatrix} = 
    \begin{pmatrix}
        g^+ & g^-\\
        g^- & g^+
    \end{pmatrix} * 
    \begin{pmatrix}
        v^+\\
        v^-
    \end{pmatrix}
    \label{eq:neg_exp}
\end{equation}

where all components $i$, $g$, and $v$ are positive. $g^+ - g^- = m$ represents each matrix element $m$ if the input voltage $v = v^+ - v^-$. The result can be computed from the output currents like $i = i^+ - i^-$. \par
Diagonalization of the $\mathbf{A}$ matrix in the S4D kernel may result in complex-valued entries. To represent these using the real-valued conductances of the mCBA, each complex number is mapped to a two-dimensional real-valued vector. When incorporating the representation of signed values as described in Equation \ref{eq:neg_exp}, the multiplication of complex-valued inputs and matrix elements expands accordingly to:
\begin{equation}
    \underline{i} * \underline{m} = 
    \begin{pmatrix}
        m_r & -m_i \\
        m_i & m_r
    \end{pmatrix} *
    \begin{pmatrix}
        i_r \\
        i_i
    \end{pmatrix} =
    \begin{pmatrix}
        g_r^+ & g_r^- & g_i^- & g_i^+ \\
        g_r^- & g_r^+ & g_i^+ & g_i^- \\
        g_i^+ & g_i^- & g_r^+ & g_r^- \\
        g_i^- & g_i^+ & g_r^- & g_r^+ 
    \end{pmatrix} * 
    \begin{pmatrix}
        v_r^+\\
        v_r^-\\
        v_i^+\\
        v_i^-
    \end{pmatrix} = 
    \begin{pmatrix}
        i_r^+\\
        i_r^-\\
        i_i^+\\
        i_i^-
    \end{pmatrix}
\end{equation}
Each diagonal matrix element is thus represented by a 4×4 block, resulting in a block-diagonal matrix structure. Consequently, the dimensions of the mCBA must be at least four times the state dimension $\mathbf{N}$.

\subsubsection{Programming quantized values on memristive crossbar}
\label{sec:mcba}
To accurately map RNN kernel matrices onto memristive conductance levels and efficiently adapt to new weight values, we employ an advanced tuning technique \citep{PPO2025AICAS} for memristive devices. This approach leverages a compact 1T1R device model specifically tailored for memristive crossbar arrays \citep{ICONS}. The model predicts optimal SET/RESET pulse parameters by considering both the deviation from the target conductance and device variability. Compared to the widely used write-and-verify approach, such as the incremental step pulse method, this technique achieves a 40\%–100\% reduction in programming time \citep{ICONS,yu2024ouroboros}.

\bibliography{literature.bib}
\bibliographystyle{bibstyle.bst}

\end{document}